\documentclass[10pt,twocolumn,letterpaper]{article}

\usepackage[pagenumbers]{cvpr}      
\definecolor{cvprblue}{rgb}{0.21,0.49,0.74}
\usepackage[pagebackref,breaklinks,colorlinks,allcolors=cvprblue]{hyperref}
\usepackage{multirow}
\usepackage{pifont}
\usepackage{makecell}
\usepackage{color}
\usepackage{dsfont}
\usepackage{bm}
\renewcommand\arraystretch{1.3}

\usepackage{makecell}
\usepackage{subcaption}
\usepackage{engord}
\usepackage{algorithm}
\usepackage{algorithmic}
\usepackage{amsmath}
\usepackage{amssymb}
\usepackage{pifont}
\usepackage{cuted}
\def\paperID{33800} 
\def\confName{CVPR}
\def\confYear{2026}

\title{
    BiEvLight: Bi-level Learning of Task-Aware Event Refinement \\
    for Low-Light Image Enhancement
    \vspace{-15pt}
}
\author{
    Zishu Yao\textsuperscript{\rm 1}, Xiang-Xiang Su\textsuperscript{\rm 1}, Shengning Zhou\textsuperscript{\rm 2}, Guang-Yong Chen\textsuperscript{\rm 1}, Guodong Fan\textsuperscript{\rm 2}\thanks{Corresponding author.}, Xing Chen\textsuperscript{\rm 1} \\
     \textsuperscript{\rm 1} College of Computer and Data Science, Fuzhou University, Fuzhou 350108, China\\
     \textsuperscript{\rm 2} College of Computer Science and Technology, Shandong Technology and Business University, \\
     Yantai 264003, China\\ 
    {\tt\small  zishuyao98@gmail.com, sxxdyx0619@163.com, 2023420043@sdtbu.edu.cn, cgykeda@mail.ustc.edu.cn,}\\ {\tt\small fgd96@outlook.com, chenxing@fzu.edu.cn}
}
\begin{document}
\maketitle
\setlength{\stripsep}{-20pt}
\begin{strip}
    \vspace{-5pt}
    \centering
    \includegraphics[width=0.99\textwidth]{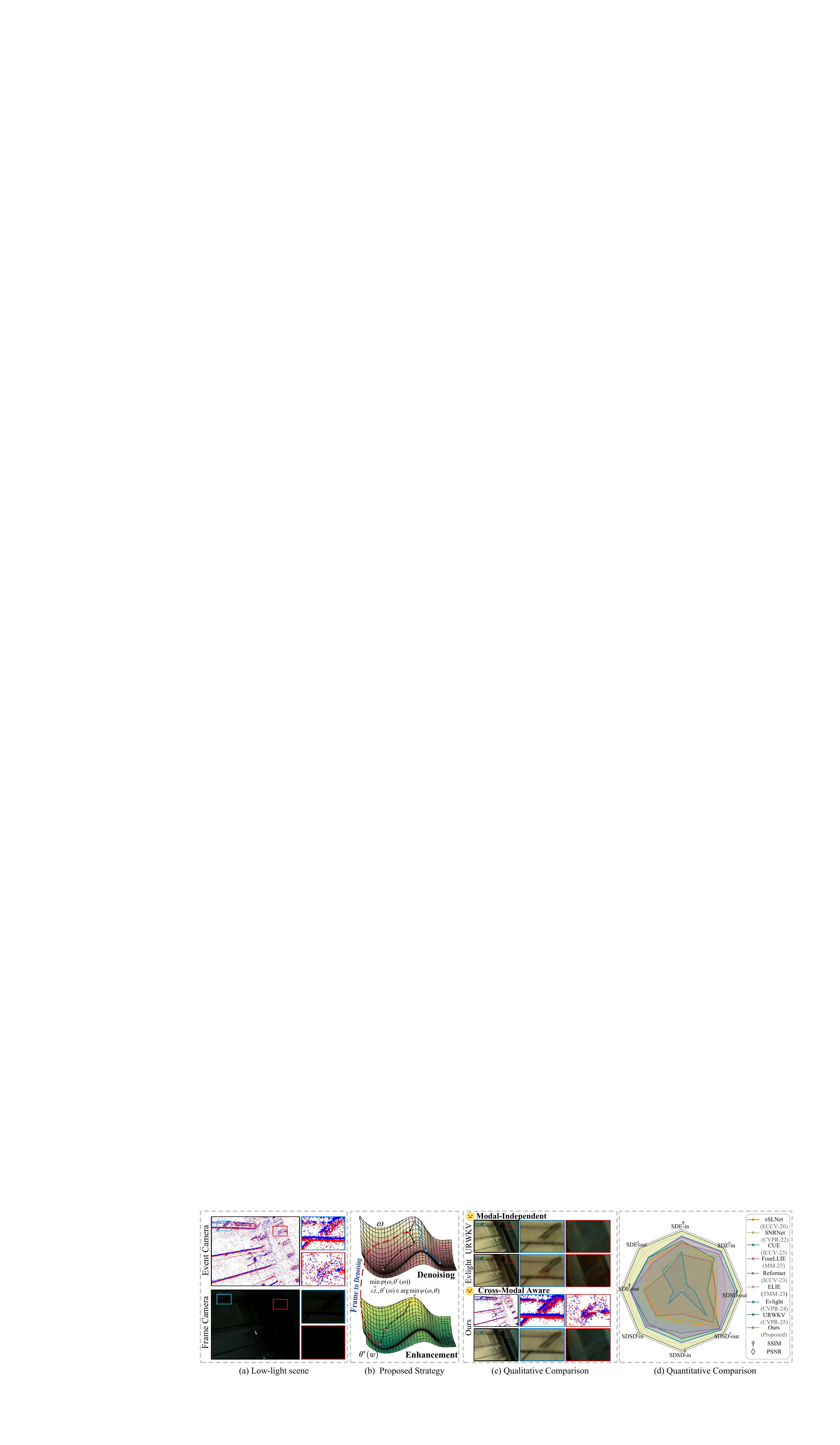}
    \captionof{figure}{(a) Two modalities exhibiting different types of degradation in low-light environments.  (b) The proposed strategy, we aim to utilize cross-modal interaction to guide denoising to achieve the event representation adapted to lower-level enhancement tasks. (c) Visual comparison with the current most representative methods, as well as visualization of events after denoising. (d) Quantitative and visual comparison with current mainstream methods, the proposed method achieves optimal results on mainstream datasets.}
    \label{fig1}
    \vspace{35pt}
\end{strip}

\begin{abstract}

Event cameras, with their high dynamic range, show great promise for Low-light Image Enhancement (LLIE). Existing works primarily focus on designing effective modal fusion strategies. However, a key challenge is the dual degradation from intrinsic background activity (BA) noise in events and low signal-to-noise ratio (SNR) in images, which causes severe noise coupling during modal fusion, creating a critical performance bottleneck. We therefore posit that precise event denoising is the prerequisite to unlocking the full potential of event-based fusion. To this end, we propose BiEvLight, a hierarchical and task-aware framework that collaboratively optimizes enhancement and denoising by exploiting their intrinsic interdependence. Specifically, BiEvLight exploits the strong gradient correlation between images and events to build a gradient-guided event denoising prior that alleviates insufficient denoising in heavily noisy regions. Moreover, instead of treating event denoising as a static pre-processing stage—which inevitably incurs a trade-off between over- and under-denoising and cannot adapt to the requirements of a specific enhancement objective—we recast it as a bilevel optimization problem constrained by the enhancement task. Through cross-task interaction, the upper-level denoising problem learns event representations tailored to the lower-level enhancement objective, thereby substantially improving overall enhancement quality.
Extensive experiments on the Real-world noise Dataset SDE demonstrate that our method significantly outperforms state-of-the-art (SOTA) approaches, with average improvements of 1.30dB in PSNR , 2.03dB in PSNR* and 0.047 in SSIM, respectively. The code will be publicly available at \url{https://github.com/iijjlk/BiEvlight}.
\end{abstract}

\vspace{-0.5cm}
\section{Introduction}
\label{sec1}

Acquiring high-quality images in nighttime dynamic scenes is crucial for applications that rely on visual content, such as autonomous driving and video surveillance. Traditional frame-based cameras typically employ long exposure times to capture images under low-light conditions. However, this approach is not only highly prone to motion blur but also introduces substantial noise. Although existing LLIE methods \cite{ma2025learning,xu2025urwkv,yang2023implicit,zheng2023empowering,cai2023retinexformer,wang2023fourllie,xu2022snr} have achieved remarkable progress in image restoration, the loss of key structural information remains largely unresolved. Consequently, achieving high-quality imaging in dynamic nighttime scenes with conventional frame-based cameras remains a significant and unresolved challenge.

Recently, the emergence of event cameras \cite{mead2023neuromorphic,brandli2014240,gallego2020event,liang2023coherent} has offered a new paradigm for addressing this challenge. Unlike traditional cameras, event cameras asynchronously record pixel-level brightness changes. Leveraging their microsecond-level temporal resolution and high dynamic range characteristics, they effectively overcome the motion blur and exposure issues inherent to traditional frame-based cameras in low-light and high-speed dynamic scenes. Building on these significant advantages of event cameras, existing Event-based LLIE methods \cite{sun2025low,jiang2023event,liang2024towards,wang2020event}  are widely dedicated to designing various fusion strategies to achieve effective integration of the event modality with image frames. 

However, existing work largely overlooks event data degradation in low-light. This stems from intrinsic sensor properties: random fluctuations in internal circuitry and dark current trigger spurious events (i.e., BA noise) when exceeding a preset threshold. Crucially, in low light, lowering the response threshold to capture faint changes dramatically amplifies BA noise \cite{guo2022low}. As Fig. \ref{fig1} (a) shows, event streams in real low-light scenes are heavily contaminated by BA noise, severely undermining events’ reliability as high-frequency priors, thus limiting enhancement performance. Current event denoising methods largely rely on local spatio-temporal correlations (e.g., nearest-neighbor filtering \cite{delbruck2008frame,liu2015design,khodamoradi2018n}, time surfaces \cite{mueggler2015lifetime,liu2024seeing}) and can effectively suppress isolated noise at low-to-moderate densities. However, in extremely low-light scenarios where high-density BA noise occurs, single-modality denoising fails to effectively suppress noise and preserve structural details. 

To alleviate the insufficient denoising in heavily noisy regions caused by the lack of modal guidance, we exploit the strong gradient correlation between images and events to develop a gradient-guided denoising prior. This prior implicitly guides the event stream denoising, effectively filtering artifacts while preserving true signals. Moreover, existing event–image fusion methods primarily focus on designing fusion modules for merging data or features from different modalities, while overlooking the intrinsic coupling between denoising and enhancement. As a result, event denoising is typically treated as a static pre-processing step prior to enhancement. Without cross-task feedback or information exchange, the resulting denoised event modality is essentially fixed and cannot adapt to the requirements of the specific enhancement objective, leading to a dilemma: excessive denoising removes critical structural details, whereas insufficient denoising allows residual noise to propagate into the fusion stage, ultimately degrading enhancement quality.

To address this challenge, we introduce BiEvLight, a task-aware bilevel learning framework. In this framework, denoising is no longer treated as a static pre-processing step but is reformulated as a bilevel optimization problem constrained by the enhancement task. Specifically, during the enhancement process that fuses complementary event-modal information, the lower-level enhancement task dynamically calibrates the upper-level event denoising through its gain feedback. The refined event signals, in turn, deliver high-SNR complementary cues to the enhancement branch. This task-aware bilevel learning mechanism adaptively achieves a delicate balance between noise suppression and structural fidelity, enabling the model to learn event representations tailored to low-light enhancement and thereby substantially improve overall enhancement quality. Our contributions can be summarized in three aspects:

\begin{itemize}
  \item We proposed BiEvLight, a task-aware bilevel learning framework that, by leveraging the intrinsic interdependence between event denoising and LLIE, redefines the denoising problem as an upper-level optimization problem constrained by the lower-level enhancement task, thereby achieving collaborative optimization of denoising and enhancement.
  \item We proposed a Spatially-adaptive Gradient-guided Denoising strategy, which leverages the strong gradient correlation between images and events to implicitly guide event stream denoising, thereby achieving precise noise suppression. 
  \item Sufficient comparisons on more benchmarks with recent advanced methods and detailed analyses are performed to prove our effectiveness.
\end{itemize}



\section{Related Works}
\label{sec:formatting}

\subsection{Low-light Image Enhancement}
\textbf{Frame-based LLIE methods:} fall into end‑to‑end \cite{fan2022multiscale,yao2024gaca,zhao2021retinexdip,  ma2023bilevel,yao2024spatial} and Retinex‑based \cite{li2018structure,wei2018deep,rahman2004retinex, jobson1997multiscale} methods. End‑to‑end methods learn a direct mapping from low‑light to normal images and do not incorporate priors on human color perception, which can lead to color distortions and reduced perceptual fidelity. Retinex‑based methods, which decompose the image $x$ as $ x=R \odot L$ and optimize illumination $L$ and reflectance $R$ separately, typically yield superior enhancement. However, in extreme photon‑starved settings where severe information loss occurs, frame‑based approaches commonly produce reconstructions with blurred or missing structural details.

\noindent\textbf{Event-based LLIE methods:} fall into only-event and event-assisted categories. Only-event methods \cite{liu2024seeing,rebecq2019high} reconstruct images solely from event streams, but the lack of image cues (e.g., initial pixel intensities) makes accurate per-pixel absolute intensity estimation difficult. Event-assisted methods use events to provide complementary edge and temporal information to frame images. For example, ELIE \cite{jiang2023event} leverages cross-modal residuals to reduce the domain gap and introduces a contrast-distribution function to lessen perceptual differences. To address the scarcity of real paired data, Liang et al. \cite{liang2024towards} released a real low-light event–image dataset SDE and proposed Evlight, which exploits differing noise distributions between events and frames and achieves strong enhancement performance. However, existing works rarely systematically model the event degradation caused by BA noise, resulting in reconstructions that remain blurred at the detail level.

\subsection{Event denoising} 

Early research primarily focused on nearest-neighbor filtering \cite{delbruck2008frame,liu2015design,khodamoradi2018n}. The core principle of these methods is to exploit the spatio-temporal locality of events, allowing only spatially adjacent or temporally close events to pass, thereby filtering out isolated, randomly generated background noise. Another class of denoising algorithms \cite{mueggler2015lifetime,liu2024seeing} treats noise events as non-ideal, instantaneous intensity change triggers. These methods achieve denoising by filtering out non-initial trigger events based on the Time Surface principle. However, research \cite{baldwin2020event} indicates that because these methods rely solely on a single modality for denoising, they often suffer from excessive residual noise, particularly in regions with severe noise.

\section{Method}
\begin{figure*}[!t]
    \centering
    \includegraphics[width=0.95\linewidth]{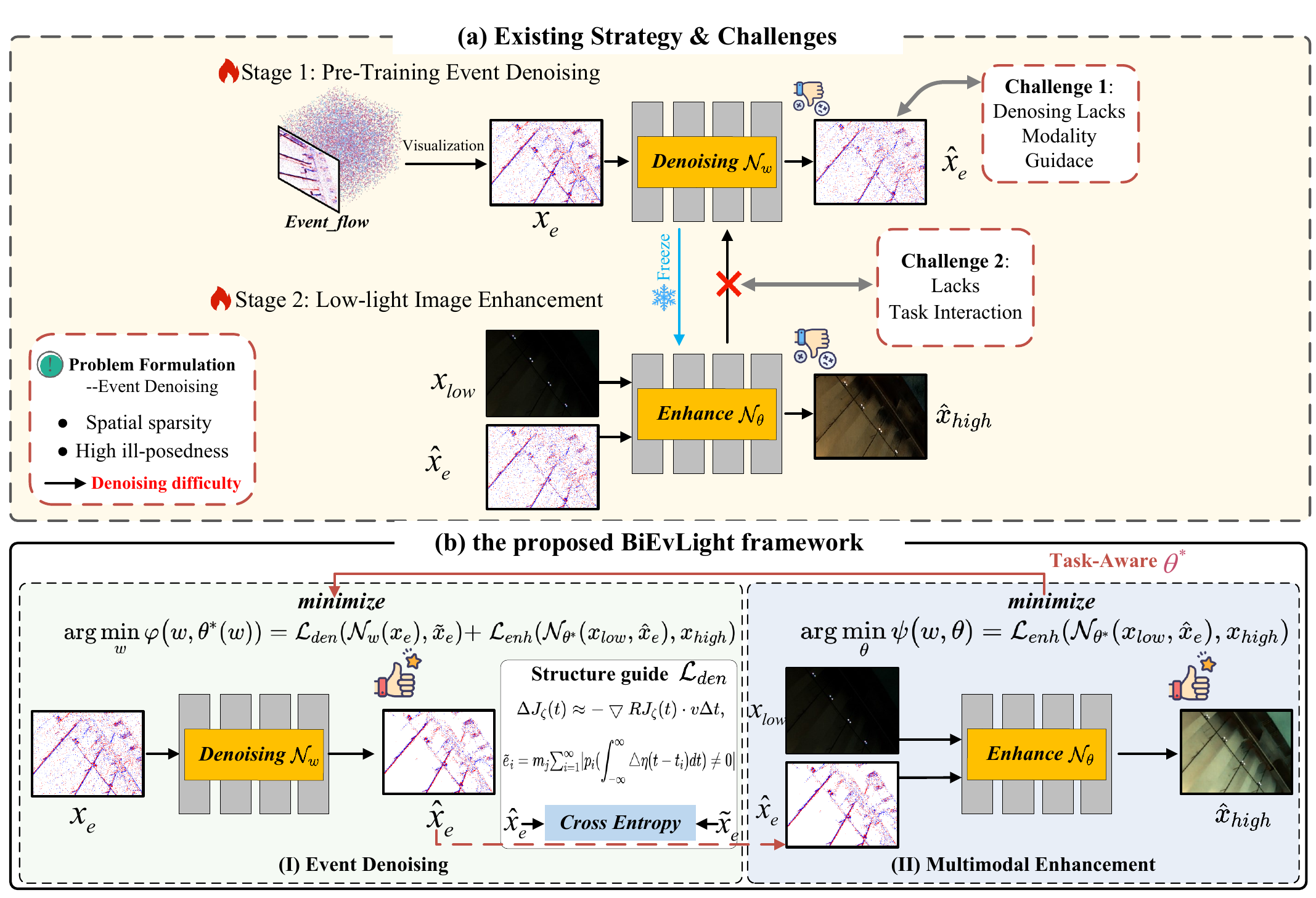}
    \caption{(a) Problems and challenges of existing methods, (b) The proposed BiEvLight framework.}
    \label{fig2}
    \vspace{-10pt}
\end{figure*}

\subsection{Problem Formulation}

\subsubsection{Mechanism of Imaging via BA Noise}
In event cameras,  $x_e = \{e_k\}^N_{k=1}$  represents the collection of events within a certain period, where a four-tuple event $ e_k=\left(i_k,t_k,p_k \right),$ is triggered when the photocurrent change at a pixel exceeds a preset threshold: 
\begin{equation}
p_k = \left\{\begin{matrix}
 +1 & if \bigtriangleup \eta  >  \epsilon  \\
 -1 & if \bigtriangleup \eta  < \epsilon  
\end{matrix}\right. ,
\end{equation}
where $i_k=(x_k,y_k)$ indicates a spatial position, $\epsilon$ is the contrast threshold, $\bigtriangleup \eta =log \left( \frac{P(i_{k})+b}{P(i_{k-1})+b}\right)$, $P(i_k)$ repersents the photocurrent of $i_k$ at $t_k$, $b$ is an infinitesimal positive number to prevent $log(0)$.

In low-contrast scenes, the contrast threshold $\epsilon$ is typically lowered to enhance sensitivity to subtle luminosity changes. However, this makes the sensor highly susceptible to internal random fluctuations, causing a large number of BA noise to be erroneously triggered. As a result, these noise events dominate the event stream, severely diminishing its value as a reliable high-frequency prior.

\subsubsection{Low-Light Imaging with Retinex}
Based on the Retinex model, we solve for the reflectance and illumination components by minimizing a regularized energy function:
\begin{equation}\label{eq5}
    E(x_r,x_l) = || x_{low}-x_r \odot x_l||^2_F +\alpha \Phi(x_r)+ \beta \bm\phi (x_l),
\end{equation}
where $||x_{low}-x_r\odot x_l||_F$ is the fidelity term, $\Phi(x_r)$ and $\bm\phi(x_l)$ are regularization terms, and $\alpha$ and $\beta$ are penalty coefficients. Here, $x_r$ denotes the degraded reflectance map, representing the intrinsic properties of the image, and $x_l$ represents the degraed illumination map, $x_{low}$ denotes the degraded low-light image. In extremely dark environments, insufficient photon capture cause many details in the  $x_r$ to be lost. The fidelity term becomes noise‑dominated, and conventional constraints alone cannot recover reliable structure.

\subsubsection{Bridging Events and Image}
By integrating the event stream with respect to time, we obtain the change in its logarithmic intensity:
\begin{equation}
\begin{split}
J_{\zeta}(t) &= \int_{t-\zeta}^{t}E(\tau) d\tau ,
\end{split}
\end{equation}
where $E(\tau) = \sum_{i=1}^{\infty }\Delta \eta_i \tau(\tau-t_i) d\tau$, and $\zeta >0$ is the temporal integration window length, $t_i$ denotes the timestamp of the $i-th$ event, $E(\tau)$ represents the continuous-time event stream. Next, the event increment within $\Delta t$ is defined as $\Delta J_{\zeta}(t)=J_{\zeta}(t)-J_{\zeta}(t-\Delta t)$, which can be approximated by 
\begin{equation}
    \Delta J_{\zeta}(t) \approx  \frac{\partial }{\partial t}  J_{\zeta}(t)\Delta t.  
\end{equation}

Based on the brightness‑constancy assumption \cite{gallego2015event}, during the time interval $\Delta t$, we have:
\begin{equation}
     \frac{\partial }{\partial t}  J_{\zeta}(t)  + \bigtriangledown x J_{\zeta}(t) \cdot v=0,
\end{equation}
where $\bigtriangledown x J_{\zeta}(t)$ is the spatial gradient, and $v$ represents the motion field.  By the Retinex theory \cite{jobson1997multiscale}, $\bigtriangledown x J_{\zeta}(t) \approx \bigtriangledown R J_{\zeta}(t)$, therefore, we obtain:
\begin{equation}\label{eq6}
    \Delta J_{\zeta}(t) \approx - \bigtriangledown R J_{\zeta}(t) \cdot v\Delta t,
\end{equation}
this indicates that events are produced by the motion of object edges; when the motion direction is perpendicular to the gradient ($v\perp  \bigtriangledown R J_{\zeta}(t)$), brightness changes occur and events are triggered. Thus, we utilize event structures to enhance degraded images, while taking into account the brightness constancy assumption. Eq.~\eqref{eq5} can be rewritten as :  
\begin{equation}
    E(x_r,x_l,x_e) = || x_{low}-x_r \odot x_l||^2_F +\alpha \Phi(x_r;x_e)+ \beta  \bm{\phi}(L).
\end{equation}

According to Eq.~\eqref{eq6}, events occur mainly at motion edges with large $|\nabla R|$, so the image gradient is a spatial–directional prior to denoising: true events are co‑supported by and aligned with the gradient, while background noise is unrelated and lacks spatiotemporal coherence. Thus, image‑guided event denoising is defined as a constrained projection onto the gradient prior: 
\begin{equation}
     \hat{x}_e = \arg \min_{\tilde{x}_e} d(x_e,\tilde{x}_e) \quad \text{s.t.} \quad \tilde{x}_e \in \mathcal{C}(G(i)),
\end{equation}
where $\hat x_e$ represents the predicted denoised event, the feasible set $\mathcal{C}(G(i))$ comprises only those events that lie on strong-gradient support, are compatible with the gradient and optical-flow directions, and exhibit local spatiotemporal coherence.

In summary, we model event denoising and low‑light image reconstruction as two complementary subproblems: $x_e$ provide motion‑consistent high‑frequency structure for the degraded reflectance $x_{r}$, while  $\nabla \tilde x_r$ (the gradient of the ground-truth reflectance)  supplies a spatial–directional prior to suppress event noise. 

\subsection{Overall Framework}
In this subsection, we introduce the overall architecture of the proposed BiEvlight, which comprises two core components: an Event Denoising Module (parameterized by $w$) and a Multimodal Enhancement Module (parameterized by $\theta$). Considering the encoder-decoder structure’s \cite{badrinarayanan2017segnet} powerful capabilities in capturing multi-scale contextual information and efficiently extracting image features, we have chosen it as the backbone for both sub-networks. The enhancement network $\mathcal{N}_\theta(\cdot)$ is expressed as:
\begin{equation}
\hat x_h = \mathcal{N}_\theta(x_{low},\hat x_e).
\end{equation}

Specifically, the enhancement network $\mathcal{N}_{\theta}(\cdot)$ is designed as a dual-branch architecture. The  low-light image $x_{low}$ is first decomposed by a pre-trained network \cite{wu2022uretinex} $\mathcal{D}(\cdot)$  to obtain initial illumination map $x_l$, and initial reflectance map $x_r$. These maps are then fed into their respective enhancement networks for mapping. Since the reflectance map represents the intrinsic properties of the object, event information is exclusively fed into the reflectance enhancement network to enhance image details. This process can be described as follows:
\begin{equation}
    \begin{cases}
        x_l,x_r = \mathcal{D}(x_{low})\\
        \hat x_l = \mathcal{N}_l(x_l)\\
        \hat x_r = \mathcal{N}_r(x_r,\hat x_e)
    \end{cases},
\end{equation}
where  the enhanced image is reconstructed as $\hat x_{high} = \hat x_l \odot \hat x_r$. The terms $\hat x_l$ and $\hat x_r$ denote the enhanced illumination and reflectance values, respectively.

Furthermore, the event denoising network $\mathcal{N}_{w}(\cdot)$ predicts the denoised events $\hat x_e$ based on the raw noisy events $x_e$, which serve as clean structural priors to guide image enhancement:
\begin{equation}
    \hat x_e = \mathcal{N}_{w}(x_e).
\end{equation}

\subsection{BiEvLight} 

\subsubsection{Spatially-adaptive Gradient-guided Denoising}

Existing event denoising methods primarily rely on spatiotemporal filters or temporal surface strategies. However, in low-light environments, when confronted with the dual challenges of sparse event signals and dense BA noise, such single-modality methods struggle to accurately distinguish between noise and legitimate events. This often leads to excessive residual noise and over-smoothing of structural details.  

From Eq. \ref{eq6}, events are triggered by brightness changes at object edges, and the image gradient not only indicates edge locations but also encodes edge magnitude and normal direction; therefore the event stream is highly correlated spatially with the image gradient. Based on this natural complementarity between images and events, we propose a Spatially-adaptive Gradient-guided Denoising strategy. Specifically, we use the structural information of the reflectance component $\tilde{x}_r$ (i.e., its gradient $\nabla\tilde{x}_r$) to guide denoising of the event stream $x_e$. The denoised events $\tilde{x}_e = \{\tilde e_k \}^N_{k=1}$ can be expressed as:
\begin{equation}
\tilde e_k =\begin{cases}
    & e_k, m_j {\textstyle \sum_{k=1}^{\infty }}|p_k(\int_{-\infty }^{\infty}\bigtriangleup \eta (t-t_k)dt )|\ne 0  \\\ 
    & \oslash , otherwise
\end{cases},
\end{equation}
the mask $m_j$ simply follows:
\begin{equation}
   m_j= \begin{cases}
    & \nabla\tilde{x}_{r,i},  \nabla\tilde{x}_{r,i} \notin (q-\mu,q+\mu)\\ 
    & 0 ,  otherwise 
\end{cases},
\end{equation}  
where $\mu$ represents the gradient supervision level, q denotes the spatially-adaptive threshold:
\begin{equation}
    q = \frac{1}{|W|^2} \sum_{(x,y)\in W_s}|\bigtriangledown \tilde x_r(x,y)|,
\end{equation}
where $W_s$ denotes the $s$-th sliding window of size $|W|$.

Event distributions exhibit significant spatial heterogeneity, with substantial differences in gradient characteristics between smooth and textured regions. A global masking strategy struggles to simultaneously achieve genuine event preservation and noise suppression across diverse regions. The proposed Spatially-adaptive threshold automatically adjusts denoising strength based on local gradient distributions, enabling region-specific optimization that preserves sparse events in smooth areas while suppressing noise in textured regions. 

Subsequently, we employ the denoised events $\tilde x_e $ as labels to implicitly guide the raw events $x_e$ in capturing the complex nonlinear dependencies between events and image gradients through a neural network.

\subsubsection{Bi-level Learning of Task-Aware} 

After leveraging reflectance gradients to guide event denoising, our goal is to effectively fuse the denoised events with the image to support low-light enhancement. However, existing multimodal fusion pipelines mostly focus on how to fuse, while overlooking the coupling between the denoising task and the enhancement task, and typically treat event denoising as a static pre-processing step placed before enhancement. In the absence of feedback and information exchange between the two tasks, the resulting denoised event modality is essentially fixed and cannot adapt to the requirements of the enhancement objective. This often leads to a fundamental dilemma: aggressive denoising removes critical structural details, whereas conservative denoising retains substantial residual noise that contaminates the fusion process and degrades the enhancement quality (see Fig. \ref{fig2}(a)).

To address this challenge, we propose a task-aware bilevel learning strategy, which reconstructs event denoising from an independent pre-processing stage into a bilevel optimization problem explicitly constrained by the enhancement objective, formulated as:
\begin{equation}\label{BLO}
    \begin{split}
    &\min_{w} \bm{\varphi}\big(w,\theta^*(w);\{x_{low},x_{high},x_e,\tilde{x}_e\}\big) \\ &s.t., \theta^*(w)\in \arg\min_{\theta} \bm{\psi} (w,\theta;\{x_{low},x_{high},\hat{x}_e\}),
\end{split}
\end{equation}  
the specific objectives of the upper and lower layers are
\begin{equation}\small
    \bm{\varphi} = \mathcal{L}_{den}(\mathcal{N}_{w}(x_e),\tilde{x}_e) + \mathcal{L}_{enh}\big(\mathcal{N}_{\theta^{*}}(x_{low},\hat{x}_e),x_{high}\big),
\end{equation}
\begin{equation}\small
     \bm{\psi} =\mathcal{L}_{enh}\big(\mathcal{N}_{\theta}(x_{low},\hat{x}_e),x_{high}\big).
\end{equation}

\begin{algorithm}[!t]
	\caption{Bi-level Learning of Task-Aware}\label{alg:BAL}
	\begin{algorithmic}[1]
		\REQUIRE The initial parameter $w_0$, $\theta_0$, step-sizes $\eta_{\theta}$, $\eta_{w}$, gradient supervision level $\mu$, iteration numbers $K$.
		\ENSURE The optimal parameters $w^*$, $\theta^*$.
        \FOR{$k=0:K-1$}
        \STATE \% Update the lower-level parameter.
        \STATE $\theta_{k+1} = \theta_k - \eta_{\theta}\nabla_{\theta}\bm\psi(w_k,\theta_k)$.
        \STATE Calculate the upper-level gradient by \eqref{upper-grad}-\eqref{f-d}.
        \STATE \% Update the upper-level parameter.
        \STATE $w_{k+1} = w_k - \eta_w\nabla_{w}\bm\varphi(w_k, \theta_k - \eta_{\theta}\nabla_{\theta}\bm\psi(w_k,\theta_k))$.
        \ENDFOR
	\end{algorithmic}
\end{algorithm}

\begin{table*}[!t]
\footnotesize
\setlength{\tabcolsep}{1.6mm}
\renewcommand\arraystretch{1.1}	
\centering
\caption{{Quantitative comparison with advanced methods of low-light image enhancement on SDE and SDSD dataset. The \textcolor{red}{red} and \textcolor{blue}{blue} represent the best and second-best values.}}

\begin{tabular}{|c|c||cccccc||ccc||c|}
    \hline

    \multicolumn{2}{|c||}{\multirow{2}{*}{Datasets}} &
    \multicolumn{6}{c||}{Image Only} &
    \multicolumn{4}{c|}{Image + Event} \\
    \cline{3-12}
    
    \multicolumn{2}{|c||}{} & SNRNet & FourLLIE & Reformer  & NeRCo & SCI++& URWKV & eSLNet & ELIE & EvLight & \multirow{2}{*}{BiEvLight} \\
        
    \cline{1-11}
    
    \multicolumn{2}{|c||}{Reference} &
        {\scriptsize \textit{CVPR' 22}} & {\scriptsize \textit{ACM MM'23}} &
        {\scriptsize \textit{ICCV' 23}} & {\scriptsize \textit{ICCV' 23}} & 
        {\scriptsize \textit{TPAMI'25}} &
        {\scriptsize \textit{CVPR' 25}} &
        {\scriptsize \textit{ECCV' 20}} & {\scriptsize \textit{TMM' 23}}  & \scriptsize \textit{CVPR' 24}& \\
    \hline

    \multirow{3}{*}{SDE-in} & PSNR$\uparrow$ &
       20.3482 & 19.7856 & 21.3169 & 19.4266 &12.4683   &21.5831 &
        21.3622 &  22.0414 & \textbf{\textcolor{blue}{22.1880}}&
        \textbf{\textcolor{red}{22.8680}} \\
    ~ & PSNR*$\uparrow$ &
        23.6785 & 23.2709 & 23.6502  & 23.2052& 22.3026   &22.6191&
        23.0825 & 23.6395& \textbf{\textcolor{blue}{23.6940}}&
        \textbf{\textcolor{red}{26.0023}} \\
    ~ & SSIM$\uparrow$ &
        0.6267&  0.6176  & 0.6873  & 0.5626& 0.4928 &0.7168&
        0.7023 & 0.7083  & \textbf{\textcolor{blue}{0.7189}} &
        \textbf{\textcolor{red}{0.7750}} \\
    \hline

    \multirow{3}{*}{SDE-out} & PSNR$\uparrow$ 
    & 21.5542 & 20.6323 & 22.3075 &18.4125  &  12.0277 & 22.3732&
        20.6244 & 22.2167 &\textbf{\textcolor{blue}{22.4372 }}&\textbf{\textcolor{red}{24.3599}}
         \\
    ~ & PSNR*$\uparrow$ & 23.3224
         & 22.1990 & 23.9459  & 22.2225 & 23.2990 &24.0964 &
        22.3612 & 23.6772 & \textbf{\textcolor{blue}{24.4223}}&\textbf{\textcolor{red}{26.1617}}
         \\
    ~ & SSIM$\uparrow$ & 0.6508
         & 0.6410 & 0.6981 & 0.5490 & 0.3033 &0.6880&
         0.6295& 0.7002  & \textbf{\textcolor{blue}{0.7070}}&\textbf{\textcolor{red}{0.7451}}
         \\
    \hline
        \multirow{3}{*}{SDSD-in} & PSNR$\uparrow$ &
       25.9892 & 24.2775 & 27.7175  &19.9693 &9.9750  &28.7034 & 25.8946
         & 28.3385  &\textbf{\textcolor{blue}{29.3563}} & \textbf{\textcolor{red}{30.7576}}
         \\
    ~ & PSNR*$\uparrow$ &
        26.4070 & 28.0437 & 28.2613  & 23.4650&29.3317   & 28.9094& 26.3361
         &29.6009 &\textbf{\textcolor{blue}{30.4038}} &  \textbf{\textcolor{red}{30.9992}}
         \\
    ~ & SSIM$\uparrow$ &
       0.8730  & 0.8668 & 0.9125  &0.7821 & 0.3108  &0.9158 &
        0.8961 & 0.9187 &\textbf{\textcolor{blue}{0.9250}} & \textbf{\textcolor{red}{0.9473}}
         \\
    \hline
        \multirow{3}{*}{SDSD-out} & PSNR$\uparrow$ & 23.6656
         & 23.4002 & 26.3934  &  21.7971&18.9373 & 22.0251&
         23.3921&   25.0251&\textbf{\textcolor{blue}{26.7407}} & \textbf{\textcolor{red}{27.4108}}
         \\
    ~ & PSNR*$\uparrow$ & 27.0967
         & 27.7418 & 28.6221& 27.6637 & 28.8727  &27.2018&
         26.3331& 29.1202 &\textbf{\textcolor{blue}{30.3066}} &\textbf{\textcolor{red}{30.4579}}
        \\
    ~ & SSIM$\uparrow$ & 0.8310
        &0.7944  &0.8485 & 0.7482 & 0.5776 &0.8176&
         0.8031& 0.8626 &\textbf{\textcolor{blue}{0.8673}} &
         \textbf{\textcolor{red}{0.8873}}
        \\
    \hline
\end{tabular}

\label{table1}
\end{table*}

As illustrated in Fig. \ref{fig2}(b), during the fusion of event-modality information for enhancement, the lower-level enhancement task feeds back its performance gain to calibrate the upper-level event denoising, while the refined event signals in turn provide complementary high-SNR cues to the enhancement process. This task-aware bilevel learning mechanism adaptively strikes a delicate balance between noise suppression and structural fidelity, learning event representations tailored to low-light enhancement and thereby substantially improving the final enhancement quality.

\subsubsection{Optimization Algorithm}
This section presents the optimization algorithm designed for the bilevel problem \eqref{BLO}. Given the nested structure of the bilevel formulation and the coupling relationship between the event denoising and low-light enhancement tasks, we first minimize the lower-level problem under the given upper-level variable $w$, and then compute the upper-level gradient by the chain rule
\begin{equation}
    \nabla_w \bm{\varphi}(w,\theta^*(w)) + \big(\frac{d\theta^*(w)}{dw}\big)^{\text T}\nabla_{\theta} \bm{\varphi}(w,\theta^*(w)).
\end{equation}

The Jacobian matrix $\frac{d\theta^*(w)}{dw}$ characterizes how the event denoising task influences the low-light enhancement task. This Jacobian is typically estimated using either Approximate Implicit Differentiation (AID) \cite{AID-Neumann2021closing} or Iterative Differentiation (ITD) \cite{reverse2015gradient,TRHG} methods. To avoid the computational complexity associated with high-order operations such as explicit Hessian inversion, we adopt a one-step truncated ITD strategy. Specifically, a one-step iteration of the lower-level optimization is performed to approximate its optimal solution:
\begin{equation}
    \theta^*(w_k)\approx \theta_k - \eta_{\theta} \nabla_\theta \bm{\psi}(w_k,\theta_k).
\end{equation}

By substituting this approximation into the upper-level objective, the gradient of the upper-level problem can be expressed as follows:
\begin{equation}\label{upper-grad}
    \nabla_w \bm\varphi(w_k,\theta^{\prime}) - \eta_{\theta}\nabla_{w\theta}^2\bm\psi(w_k,\theta_k)\nabla_{\theta^{\prime}}\bm\varphi(w_k,\theta^{\prime}),
\end{equation}
where $\theta^{\prime} = \theta_k - \eta_{\theta} \nabla_\theta \bm\psi(w_k,\theta_k)$. To further reduce computational overhead, the Hessian–vector product in the second term is approximated using a finite-difference scheme,
\begin{equation}\label{f-d}\small
\nabla_{w\theta}^2\bm\psi(w_k,\theta_k)\nabla_{\theta^{\prime}}\bm\varphi(w_k,\theta^{\prime})\approx \frac{\nabla_{w}\bm\psi(w_k,\theta^{+})-\nabla_{w}\bm\psi(w_k,\theta^{-})}{2\epsilon},
\end{equation}
where $\theta^{\pm}=\theta_k\pm\epsilon\nabla_{\theta}\bm\varphi(w_k,\theta^{\prime})$ and $\epsilon$ is a small scalar set to $0.01/\|\nabla_{\theta}\bm\varphi(w_k,\theta^{\prime})\|_2$, following the setting in \cite{liu2018darts}. The complete update procedure is summarized in Algorithm \ref{alg:BAL}.

\begin{figure*}[!t]
\centering
    \includegraphics[width=0.98\linewidth]{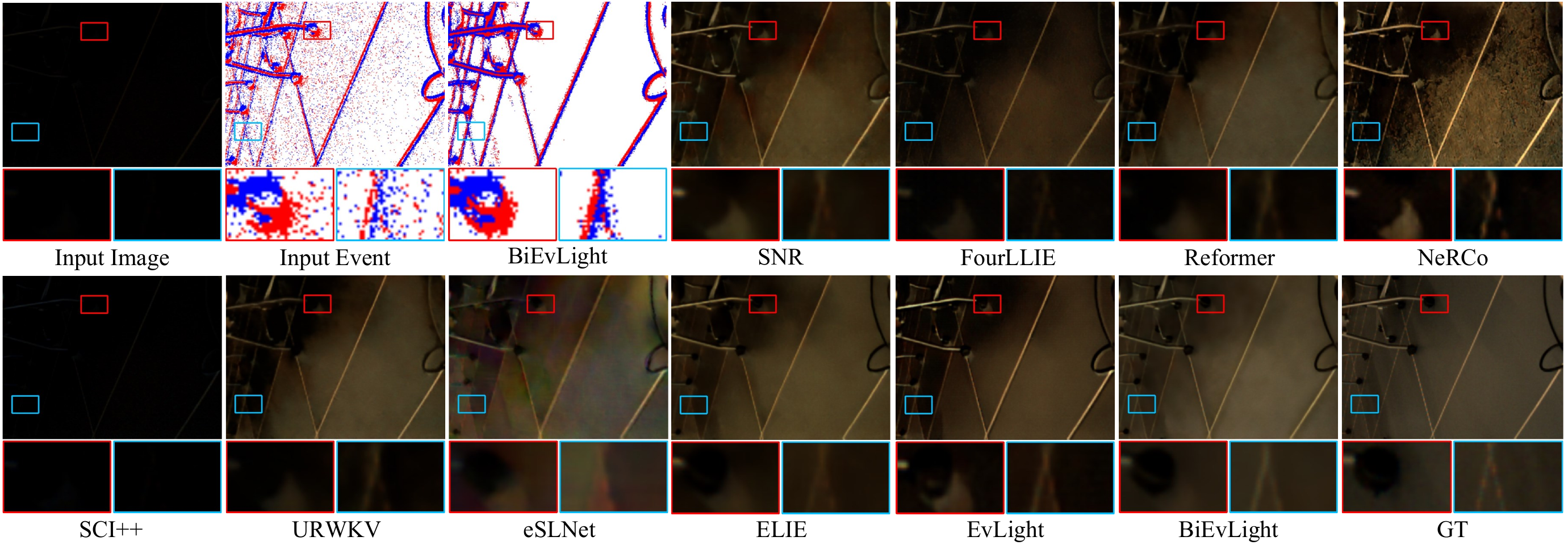}
    \caption{Qualitative results on SDE-in dataset.}
    \label{fig3}
\centering
\end{figure*}

\begin{figure*}[!t]
\centering
    \includegraphics[width=0.98\linewidth]{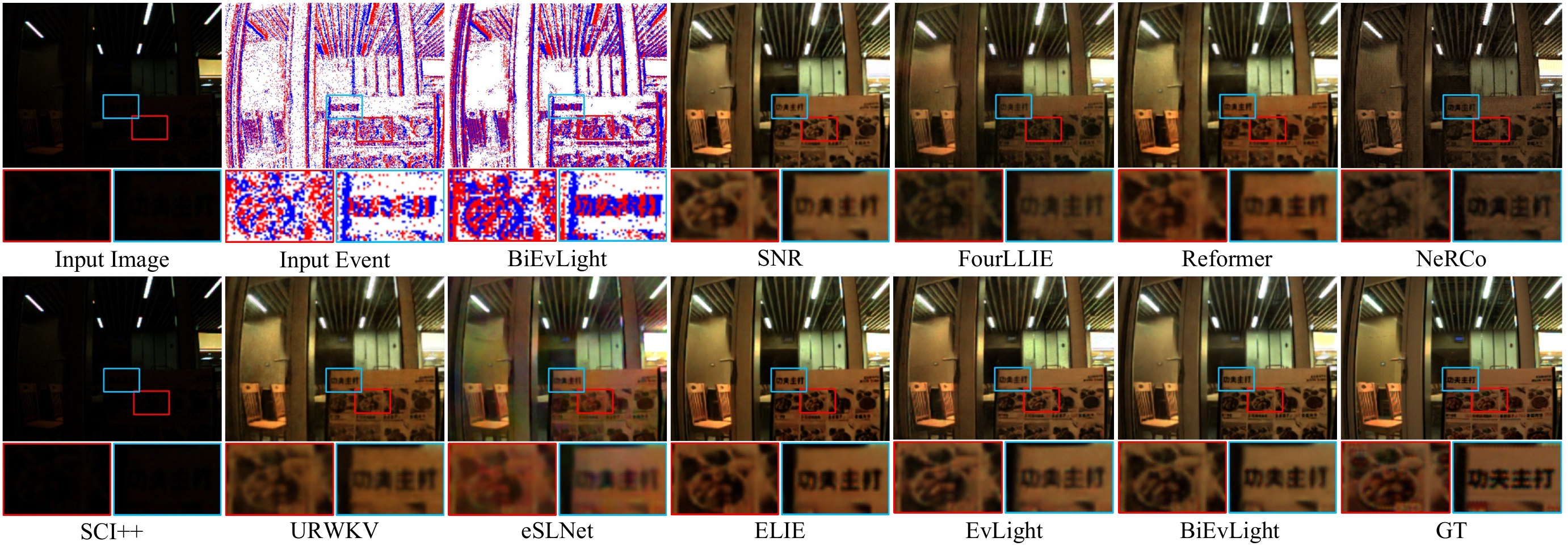}
    \caption{Qualitative results on SDE-out dataset.}
    \label{fig4}
\centering
\end{figure*}
\subsection{Loss Function}

In this paper, the loss function is primarily divided into two aspects: low-light enhancement loss and event denoising loss. Firstly, the enhancement loss can be defined as:
\begin{equation}
 \mathcal{L}_{enh} =   ||\hat x_{high} - x_{high}||_1 + \alpha ||\hat x_l - \tilde x_l||_1 + \beta ||\hat x_r-\tilde x_r||_1,
\end{equation} 
where $\alpha=0.5$, $\beta=0.5$. Meanwhile, we employ cross-entropy loss as the event denoising loss, which is defined as follows:
\begin{equation} 
\label{eq:cross_entropy_3_class}
L = -\sum_{c=1}^{3} {x^c_e} \log({\hat{x}^c_e}),
\end{equation}
where ($c \in \{1,2,3\}$) denotes the event category (i.e., positive event, negative event, and no-event region).

\vspace{-6pt}
\section{Experiments}
\subsection{Implementation Details}

    
    \noindent\textbf{Benchmarks and metrics:}
    We employ the Peak Signal-to-Noise Ratio (PSNR) \cite{hore2010image}, the Structural Similarity Index Measure (SSIM) \cite{wang2004image}, and PSNR* \cite{wu2023learning} for evaluating image restoration performance.
    We compare BiEvLight against recent SOTA methods under two distinct configurations: 1) RGB-only Input: This category includes methods that utilize only RGB images as their input modality: SNRNet~\cite{xu2022snr}, FourLLIE~\cite{wang2023fourllie}, Retinexformer~\cite{cai2023retinexformer}, NeRCo~\cite{yang2023implicit}, SCI++~\cite{ma2025learning}, and URWKV~\cite{xu2025urwkv}. 2) RGB and Paired Event Stream Input: This category comprises methods that leverage both RGB images and their corresponding event streams: eSLNet~\cite{wang2020event}, ELIE~\cite{jiang2023event}, and Evlight~\cite{liang2024towards}. For Evlight, we utilized its officially provided pre-trained weights for testing. All other comparative methods were re-trained from their publicly available codebases to ensure fair comparison.

    \noindent\textbf{Datasets:}
    This paper evaluates the proposed BiEvLight in LLIE using the SDE and SDSD datasets. SDE dataset \cite{liang2024towards} contains 91 image–event paired sequences (43 indoor sequences and 48 outdoor sequences), with 76 sequences used for training and 15 sequences for testing. SDSD dataset \cite{wang2021seeing} provides 150 paired sequences, of which 125 are used for training and 25 for testing. The original resolution is 1920 $\times$ 1080 and, following \cite{liang2024towards}, it is downsampled to 346 $\times$ 260. Noisy event streams are synthesized via an event simulator.
\vspace{-0.5cm}
\subsection{Comparison and Evaluation}

\vspace{-0.2cm}
 \textbf{Comparison on  SDE Dataset:} Quantitative results in Tab. \ref{table1} showcase our method’s superior performance on the SDE dataset. For the SDE-in and SDE-out tasks, our method surpasses the current SOTA method (Evlight) by a significant margin, achieving PSNR improvements of 0.68 dB and 1.92 dB, respectively. To further evaluate image restoration effectiveness beyond light fitting, we also computed the PSNR* metric. Our BiEvLight notably outperforms the SOTA, with PSNR* values higher by 1.58 dB for SDE-in and 1.73 dB for SDE-out. These compelling results comprehensively validate the effectiveness of our proposed BiEvLight. Qualitatively, as depicted in Fig. \ref{fig3} (for indoor scenes) and Fig. \ref{fig4} (for outdoor scenes), BiEvLight consistently reconstructs clear edges in dark areas. Furthermore, BiEvLight achieves precise event denoising. As highlighted in Fig. \ref{fig4}, while the raw events in the magnified region are severely obscured by noise, making effective information unrecognizable, the denoised events produced by BiEvLight clearly reveal text information within the scene, further demonstrating the effectiveness of our approach.

\begin{figure}[!t]
    \centering
    \includegraphics[width=1\linewidth]{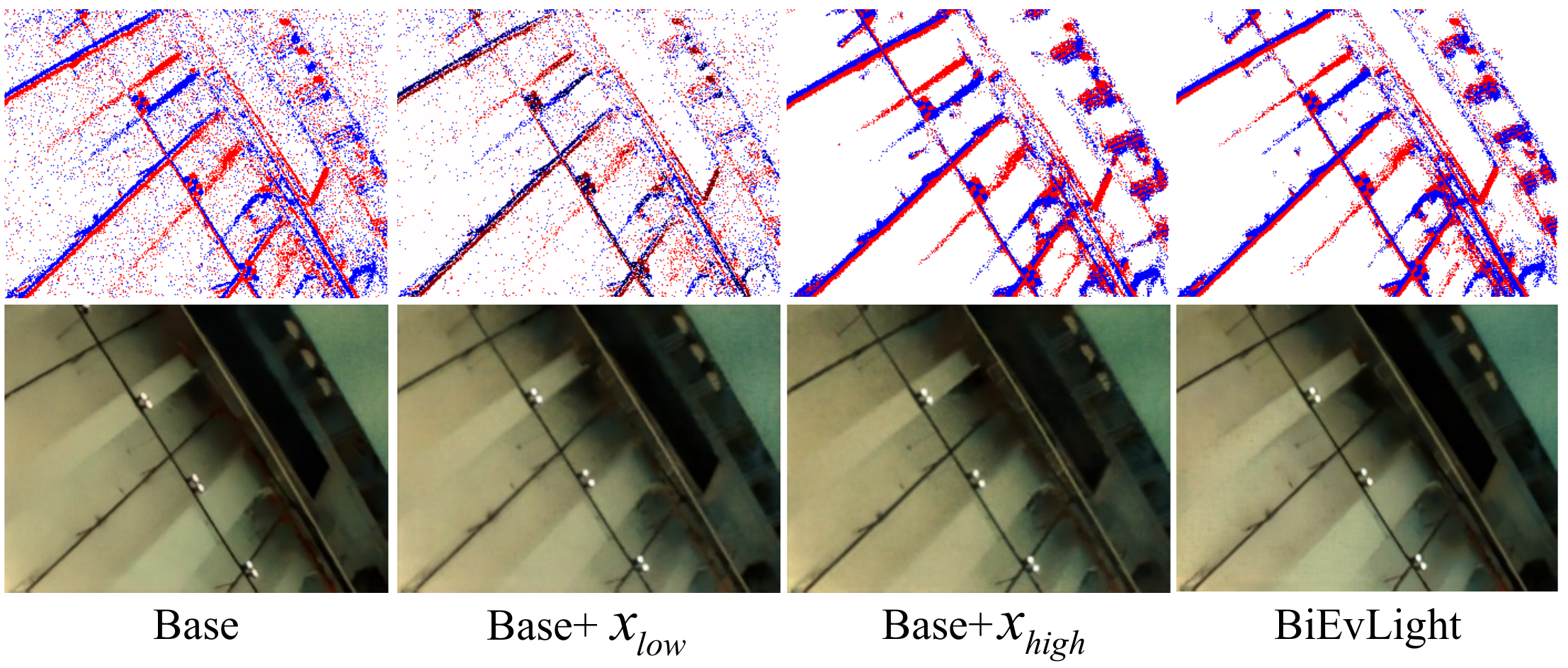}
    \caption{Visualization of Ablation Results for Event Denoising}
    \label{fig6}
\end{figure}

 \noindent\textbf{Comparison on  SDSD Dataset:}  
 Similarly, we conducted comparative experiments on the SDSD dataset, with quantitative results detailed in Tab. \ref{table1}. BiEvLight achieved optimal performance in both indoor and outdoor scenarios. Specifically, for SDSD-in and SDSD-out tasks, our method surpasses Evlight with PSNR improvements of 1.41 dB and 0.67 dB, respectively. For the PSNR* , BiEvLight demonstrates competitive advantages with improvements of 0.60 dB for SDSD-in and 0.15 dB for SDSD-out. These results comprehensively demonstrate the effectiveness of the proposed BiEvLight method.
 


\vspace{-0.2cm}
\begin{table}[!t]
\caption{Ablation study of Event denoising. }
\centering
\small
\renewcommand{\arraystretch}{1}

\setlength{\tabcolsep}{0.3mm}
\begin{tabular}{ccccccc}

				\toprule

    \multirow{2}{*}{\text{Strategy}} &
    \multicolumn{3}{c}{SDE-in} &
    \multicolumn{3}{c}{SDE-out} 
    \\
    \cline{2-7}
				   {} & \text{PSNR} & PSNR* &SSIM& \text{PSNR} & PSNR* &SSIM \\
                \midrule
                Base &21.4295 & 21.9412 & 0.7024 & 21.8932& 22.4921 &0.6941
                \\  
			    Base+$x_{low}$  &22.0431& 24.6423& 0.7423 &23.0923 &  24.8432& 0.7124
                  \\  
			      Base+$x_{high}$  &22.5398&25.5993 & 0.7532 & 23.8762& 25.6823 &0.7264
                \\
                \midrule
				BiEvLight  & \textbf{22.8680} & \textbf{26.0023} &\textbf{0.7750} & \textbf{24.3599}&\textbf{26.1617} &\textbf{0.7451}
                \\
                \bottomrule
		\end{tabular}

\label{table2}
\end{table}

\begin{table}[!t]
\caption{Ablation study of Bilevel learning.}
\centering
\footnotesize
\renewcommand{\arraystretch}{1.2}
\setlength{\tabcolsep}{0.3mm}
\begin{tabular}{ccccccc}
				\toprule

    \multirow{2}{*}{\text{Strategy}} &
    \multicolumn{3}{c}{SDE-in} &
    \multicolumn{3}{c}{SDE-out} 
    \\
    \cline{2-7}
				   {} & \text{PSNR} & PSNR* &SSIM& \text{PSNR} & PSNR* &SSIM \\
                \midrule
               Joint Learning & 21.9812 & 24.2307 & 0.7356 & 22.8156 & 24.5623 & 0.7087 \\
Alternating Learning & 22.6785 & 25.3691 & 0.7698 & 24.1234 & 25.8945 & 0.7398 \\
                \midrule
							BiEvLight  & \textbf{22.8680} & \textbf{26.0023} &\textbf{0.7750} & \textbf{24.3599}&\textbf{26.1617} &\textbf{0.7451}
                \\
                \bottomrule
		\end{tabular}

\label{table3}
\end{table}
\subsection{Ablation Studies and Analysis}
We conduct ablation experiments on the SDE dataset to evaluate the effectiveness of the proposed method.

\textbf{Impact of Events denoising:} We first verify the performance gains of event denoising on the task, and subsequently validate the improvements brought by the proposed strategy to both denoising and enhancement tasks. As shown in Tab. \ref{table2}, we refer to the network without event denoising as the Base model. The second row employs only the reflectance gradient from low-light images for denoising, while the third row utilizes the proposed Spatially-adaptive Gradient-guided Denoising strategy to learn the denoising mapping. Tab. \ref{table2} demonstrates that the restoration performance is significantly improved after applying the denoising strategy. Simultaneously, Fig. \ref{fig6} presents the qualitative comparison results, which reveal that when BiEvLight is used to guide the denoising model, the model maintains its denoising capability while preserving more features that are beneficial for enhancement.

\textbf{Impact of Bilevel learning:} We validate the gains of the proposed Bilevel-learning framework. Specifically, we optimize the two networks using joint optimization, alternating optimization, and our proposed BiEvLight; the quantitative results are shown in Tab. \ref{table3}. The quantitative results show that with joint training, concurrent optimization of the two tasks causes gradient conflicts and unstable convergence, preventing either task from being adequately refined. Alternating learning trains the denoising network first and then fixes its parameters while optimizing the enhancement network, but this paradigm lacks interaction between tasks and thus limits further performance improvement. The experimental results further confirm the effectiveness of the proposed BiEvLight.

\section{Conclusion}
This work proposed BiEvLight, a hierarchical and task-aware framework designed to address critical challenges in event-based LLIE arising from the lack of task-aware guidance and cross-modal interaction in existing methods. Specifically, we reformulate the event denoising and low-light enhancement problems into a dynamic, task-aware optimization framework. Through this approach, the enhancement model’s performance gains feed back to calibrate how events should be denoised, while the refined events in turn provide clearer structural guidance for enhancement. This bidirectional feedback mechanism achieves synergistic improvements in both event denoising and low-light enhancement. Additionally, we propose a Spatially-adaptive Gradient-guided Denoising strategy that leverages the strong gradient correlation between events and images to achieve precise event denoising. Extensive experiments demonstrate the significant advantages of our method.
\clearpage
{
    \small
    \bibliographystyle{ieeenat_fullname}
    \bibliography{main}
}

\clearpage
\setcounter{page}{1}
\maketitlesupplementary

In this supplementary material, we provide additional  details and experimental results. First, we outline the framework details of BiEvLight in Sec. \ref{sec1}. Next, Sec. \ref{sec2} offers descriptions of the training settings. In Sec. \ref{sec3}, we provides more qualitative examples to illustrate the model’s  performance. 

\begin{figure}[h]
  \centering
  \includegraphics[width=0.98\linewidth]{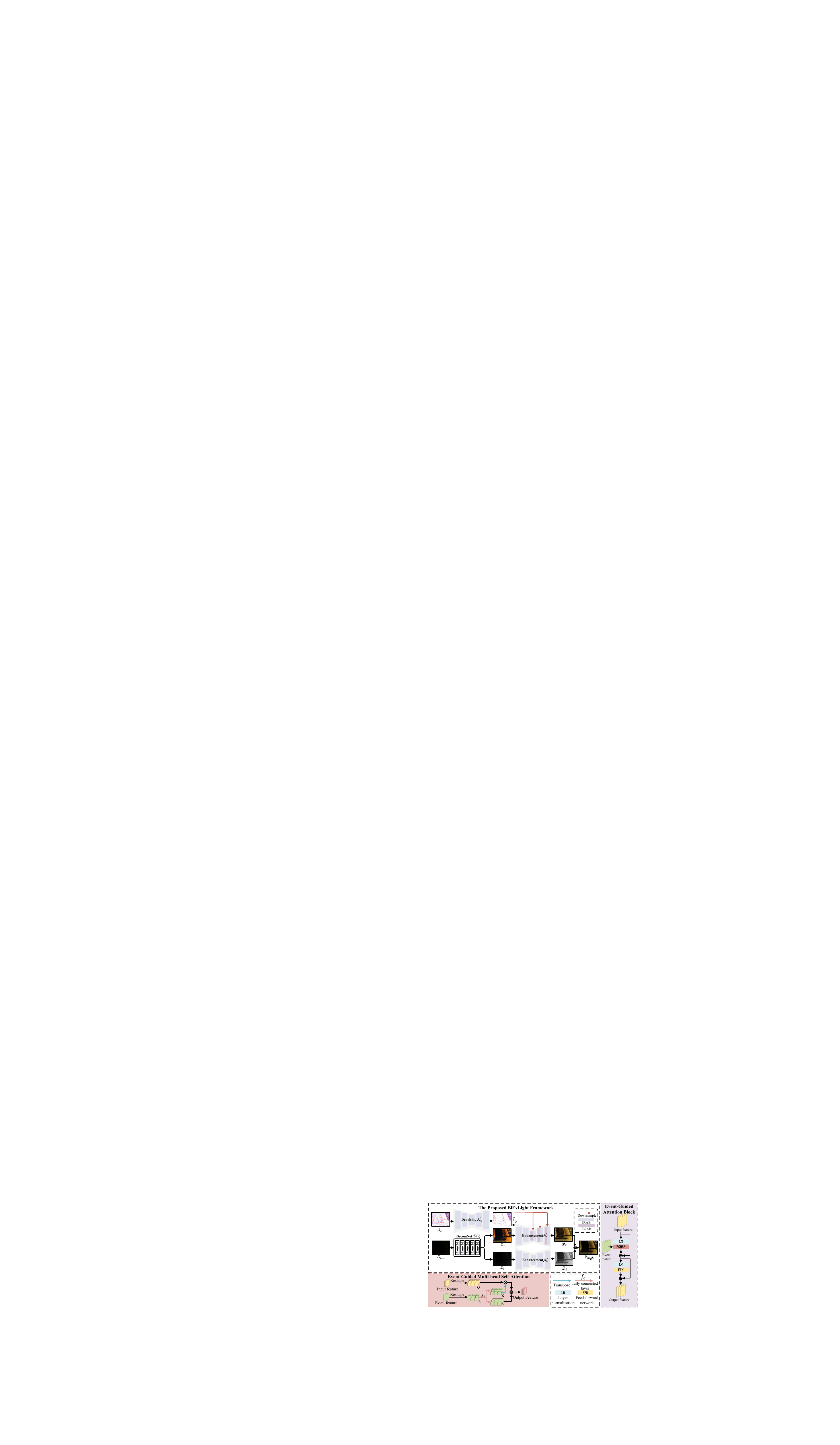}
   \vspace{-8pt}
   \caption{Qualitative analysis of denoising hyperparameters $q$ and window size $W$.}
   \label{fig8}
\end{figure}

\section{Framework Details}

\begin{table*}[!t]
\footnotesize
\setlength{\tabcolsep}{1.6mm}
\renewcommand\arraystretch{1.1}	
\centering
\caption{{Quantitative comparison with advanced methods of low-light image enhancement on SDE and SDSD dataset. The \textcolor{red}{red} and \textcolor{blue}{blue} represent the best and second-best values.}}

\begin{tabular}{|c|c||cccccc||ccc||c|}
    \hline

    \multicolumn{2}{|c||}{\multirow{2}{*}{Datasets}} &
    \multicolumn{6}{c||}{Image Only} &
    \multicolumn{4}{c|}{Image + Event} \\
    \cline{3-12}
    \multicolumn{2}{|c||}{} & SNRNet & FourLLIE & Reformer  & NeRCo & SCI++& URWKV & eSLNet & ELIE & EvLight & \multirow{2}{*}{BiEvLight} \\
        
    \cline{1-11}
    
    \multicolumn{2}{|c||}{Reference} &
        {\scriptsize \textit{CVPR' 22}} & {\scriptsize \textit{ACM MM'23}} &
        {\scriptsize \textit{ICCV' 23}} & {\scriptsize \textit{ICCV' 23}} & 
        {\scriptsize \textit{TPAMI'25}} &
        {\scriptsize \textit{CVPR' 25}} &
        {\scriptsize \textit{ECCV' 20}} & {\scriptsize \textit{TMM' 23}}  & \scriptsize \textit{CVPR' 24}& \\
    \hline

    \multirow{3}{*}{SDE-in} & PSNR$\uparrow$ &
       20.3482 & 19.7856 & 21.3169 & 19.4266 &12.4683   &21.5831 &
        21.3622 &  22.0414 & \textbf{\textcolor{blue}{22.1880}}&
        \textbf{\textcolor{red}{22.8680}} \\
    ~ & PSNR*$\uparrow$ &
        23.6785 & 23.2709 & 23.6502  & 23.2052& 22.3026   &22.6191&
        23.0825 & 23.6395& \textbf{\textcolor{blue}{23.6940}}&
        \textbf{\textcolor{red}{26.0023}} \\
    ~ & SSIM$\uparrow$ &
        0.6267&  0.6176  & 0.6873  & 0.5626& 0.4928 &0.7168&
        0.7023 & 0.7083  & \textbf{\textcolor{blue}{0.7189}} &
        \textbf{\textcolor{red}{0.7750}} \\
    \hline

    \multirow{3}{*}{SDE-out} & PSNR$\uparrow$ 
    & 21.5542 & 20.6323 & 22.3075 &18.4125  &  12.0277 & 22.3732&
        20.6244 & 22.2167 &\textbf{\textcolor{blue}{22.4372 }}&\textbf{\textcolor{red}{24.3599}}
         \\
    ~ & PSNR*$\uparrow$ & 23.3224
         & 22.1990 & 23.9459  & 22.2225 & 23.2990 &24.0964 &
        22.3612 & 23.6772 & \textbf{\textcolor{blue}{24.4223}}&\textbf{\textcolor{red}{26.1617}}
         \\
    ~ & SSIM$\uparrow$ & 0.6508
         & 0.6410 & 0.6981 & 0.5490 & 0.3033 &0.6880&
         0.6295& 0.7002  & \textbf{\textcolor{blue}{0.7070}}&\textbf{\textcolor{red}{0.7451}}
         \\
    \hline
        \multirow{3}{*}{SDSD-in} & PSNR$\uparrow$ &
       25.9892 & 24.2775 & 27.7175  &19.9693 &9.9750  &28.7034 & 25.8946
         & 28.3385  &\textbf{\textcolor{blue}{29.3563}} & \textbf{\textcolor{red}{30.7576}}
         \\
    ~ & PSNR*$\uparrow$ &
        26.4070 & 28.0437 & 28.2613  & 23.4650&29.3317   & 28.9094& 26.3361
         &29.6009 &\textbf{\textcolor{blue}{30.4038}} &  \textbf{\textcolor{red}{30.9992}}
         \\
    ~ & SSIM$\uparrow$ &
       0.8730  & 0.8668 & 0.9125  &0.7821 & 0.3108  &0.9158 &
        0.8961 & 0.9187 &\textbf{\textcolor{blue}{0.9250}} & \textbf{\textcolor{red}{0.9473}}
         \\
    \hline
        \multirow{3}{*}{SDSD-out} & PSNR$\uparrow$ & 23.6656
         & 23.4002 & 26.3934  &  21.7971&18.9373 & 22.0251&
         23.3921&   25.0251&\textbf{\textcolor{blue}{26.7407}} & \textbf{\textcolor{red}{27.4108}}
         \\
    ~ & PSNR*$\uparrow$ & 27.0967
         & 27.7418 & 28.6221& 27.6637 & 28.8727  &27.2018&
         26.3331& 29.1202 &\textbf{\textcolor{blue}{30.3066}} &\textbf{\textcolor{red}{30.4579}}
        \\
    ~ & SSIM$\uparrow$ & 0.8310
        &0.7944  &0.8485 & 0.7482 & 0.5776 &0.8176&
         0.8031& 0.8626 &\textbf{\textcolor{blue}{0.8673}} &
         \textbf{\textcolor{red}{0.8873}}
        \\
    \hline
\end{tabular}

\label{table1}
\end{table*}

The training strategy of the proposed BiEvLight has been elaborated in the main paper. Here, we provide a more detailed description of the network composition of BiEvLight and the processing of input images $x_{low}$ and denoising event $\hat{x}_e$. 

Specifically, as shown in Fig. \ref{fig8}, BiEvLight consists of an event denoising network $\mathcal{N}_{w}(\cdot)$ and an enhancement network $\mathcal{N}_{\theta}(\cdot)$ (comprising three subnetworks: a pre-trained decomposition network \cite{wu2022uretinex} $\mathcal{D}(\cdot)$, a reflectance enhancement network $\mathcal{N}_{r}(\cdot)$, and an illumination enhancement network $\mathcal{N}_{l}(\cdot)$, where $\mathcal{D}(\cdot)$ is frozen and excluded from the training process).

Regarding the detailed network configuration, excluding the pre-trained decomposition network $\mathcal{D}(\cdot)$, the remaining subnetworks are built upon a unified 3-level encoder-decoder architecture. Specifically, in the reflectance enhancement network $\mathcal{N}_r(\cdot)$, the denoised event stream $\hat{x}_e$ is first encoded via convolution and downsampling operations to generate multi-scale features $F_{ev}$ that align with the spatial resolution of the image features $F_{img}$. Subsequently, these event features are hierarchically injected into the network starting from the bottleneck layer via the Event-Image Feature Fusion Block, thereby incorporating multi-scale structure-guided information.

Specifically, the Event-Image Attention Block is designed to efficiently incorporate structural priors from the event modality into image features via a cross-attention mechanism. In the core attention computation, we project the input image features as Queries ($Q$
) and the event features as Keys ($K$) and Values ($V$). To reduce computational complexity while preserving a global receptive field, we adopt a Transposed Attention strategy, calculating the feature cross-covariance matrix $A$ in the channel dimension rather than the traditional spatial dimension. Furthermore, an implicit positional encoding based on depth-wise separable convolutions is introduced in parallel to supplement local spatial context, followed by a residual connection for effective feature aggregation:
\begin{equation}
\begin{aligned}
\mathbf{Q} &= \mathbf{W}_q \mathbf{F}_{img}, \quad \mathbf{K} = \mathbf{W}_k \mathbf{F}_{ev}, \quad \mathbf{V} = \mathbf{W}_v \mathbf{F}_{ev} ,\\
\mathbf{A} &= \text{Softmax}\left( \alpha \cdot \bar{\mathbf{K}}\bar{\mathbf{Q}}^\top \right), \\
\mathbf{F}_{out} &= \mathbf{W}_p \left( \mathbf{A}\mathbf{V} \right) + \text{CPE}(\mathbf{V}) + \mathbf{F}_{img},
\end{aligned}
\label{eq:eiab}
\end{equation}
where $\mathbf{W}_{q}, \mathbf{W}_{k}, \mathbf{W}_{v}, \mathbf{W}_{p}$ denote the linear projections, and $\bar{\mathbf{Q}}, \bar{\mathbf{K}}$ represent the $L_2$-normalized query and key features, respectively. $\alpha$ is a learnable scaling factor used to modulate the attention map $\mathbf{A}$, which is calculated along the channel dimension. Additionally, $\text{CPE}(\cdot)$ refers to the conditional positional encoding implemented by depth-wise separable convolutions to capture local spatial context.

Given the input image $x_{low}$ and the denoised events $\hat{x}_e$, we first employ shallow projection layers to process the distinct modal information independently. Subsequently, the shallow features from both modalities are fused and propagated forward. The merged features pass through three encoder stages, where they undergo interactive fusion with the aforementioned event features, followed by three decoder stages and an output projection layer to generate the enhanced reflectance $\hat{x}_r$. Finally, based on Retinex theory \cite{jobson1997multiscale}, the enhanced image is obtained as $\hat{x}_{high} = \hat{x}_{r} \odot \hat{x}_{l}$.

\section{Complexity and hyperparameters Analysis}

 As shown in Tab. \ref{tableab}, BiEvLight introduces no extra inference latency. Despite a ~23\% rise in training time over the alternating strategy, it yields a notable \textbf{0.64 dB} gain in PSNR*. Notably, the proposed framework is compatible with single-loop bilevel algorithms, which can be leveraged for scaling to larger networks.
 
\begin{table}[h]
    \caption{Comparison of model parameters (M), FLOPs (G) and FPS (Frames) on $256 \times 256$ images, training time (Hours), and PSNR* on SED\_in Dataset.}
    \vspace{-10pt}
    \centering
    \footnotesize
    \renewcommand{\arraystretch}{0.8}

    \setlength{\tabcolsep}{3pt} 
    \begin{tabular}{cccccc|c}
        \toprule

        \textbf{Methods} & \textbf{SSIM}&\textbf{PSNR*} &\textbf{\#Params } & \textbf{FLOPs} & \textbf{FPS} & \textbf{Train Cost }  \\
        \midrule
        URWKV  &0.716 & 22.61 & 2.25    &  36.67   &  11 & 56 \\
        eSLENet&0.702 & 23.08 & 0.19    & 471.81     &  11 & 28 \\
        ELIE  &0.708  & 23.63 &220.01 & 201.57 & 4 & 93 \\
        EvLight& 0.718& 23.69 &22.73 & 194.24 & 18 & 34 \\
        \midrule
        Joint& 0.735& 24.23 &2.471 &61.58 & 24& 19 \\
        Alternating & 0.769& 25.36 &2.471 &61.59 & 24 & 39
        \\
        \midrule
        BiEvLight & 0.775 & 26.00  & 2.471 & 61.59  & 24 &  48\\
        \bottomrule
        
    \end{tabular}
    \label{tableab}
\end{table}
 Similar to common practice, the \textbf{step size} in BiEvLight follows a cosine annealing schedule with restarts. It is initialized at \textbf{2e-4} and decays to \textbf{1e-6} over \textbf{150k} iterations. We observed no instability with this setup. Regarding the finite-difference scale $\epsilon$, it is \textbf{adaptively scaled} as $\epsilon = m / ||\nabla_{\theta} \varphi(w_k,\theta^{\prime})\|_2 $ to ensure numerical stability during optimization. Robustness tests (Tab. \ref{tableab2}) show stable performance across different *m* (we used \textbf{0.01}).
\begin{table}[h]
    \caption{Sensitivity analysis of the scaling factor $m$ on the SED\_in dataset.}
    \vspace{-20pt}
    \centering
    \footnotesize
    \renewcommand{\arraystretch}{0.8}

    \setlength{\tabcolsep}{15pt} 
    \begin{tabular}{ccc}
        \toprule

        \textbf{m} & \textbf{SSIM} & \textbf{PSNR*} \\
        \midrule
        0.005   &  0.772    &   25.85   \\
        0.01 &    0.775   & 26.00      \\
        0.05    &  0.771 & 25.92 \\
        0.1 &  0.772 &  25.76\\
        \bottomrule
        
    \end{tabular}
    \label{tableab2}
\end{table}

\begin{figure}[h]
  \centering
  \includegraphics[width=0.9\linewidth]{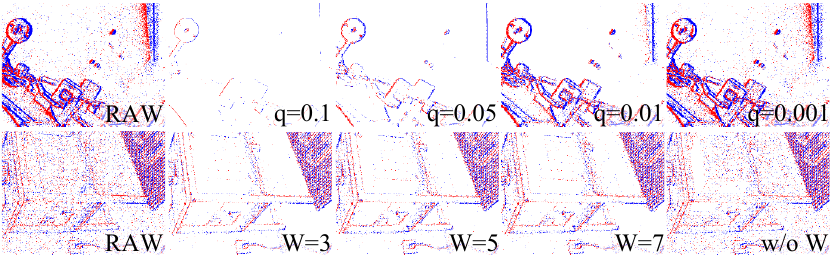}
  \vspace{-10pt}
   \caption{The network architecture of the proposed BiEvLight.}
   \label{fig9}
\end{figure}

For the sensitivity of threshold $q$ and denoising window $W$ (Fig. \ref{fig9}),
in BiEvLight, we set $ \bm {q=0.01}$ and $\bm {W=5}$. As shown, high $q$ erases details, while low $q$ retains noise. We chose $q=0.01$ as a trade-off. Although slight texture loss or noise may persist, this is specifically resolved by our interaction strategy, which uses feedback to recover textures and suppress noise. The second row illustrates the impact of varying $W$ with $q=0.01$. The denoising results remain consistent across different window sizes, demonstrating the robustness of our strategy. The results without the local window strategy (w/o $W$) are also displayed. Furthermore, the denoising pseudo-labels are guided by the gradient of the ground truth reflectance, thereby avoiding inaccurate gradient estimations caused by low-light degradation.

\begin{figure*}[!t]
\centering
    \includegraphics[width=0.98\linewidth]{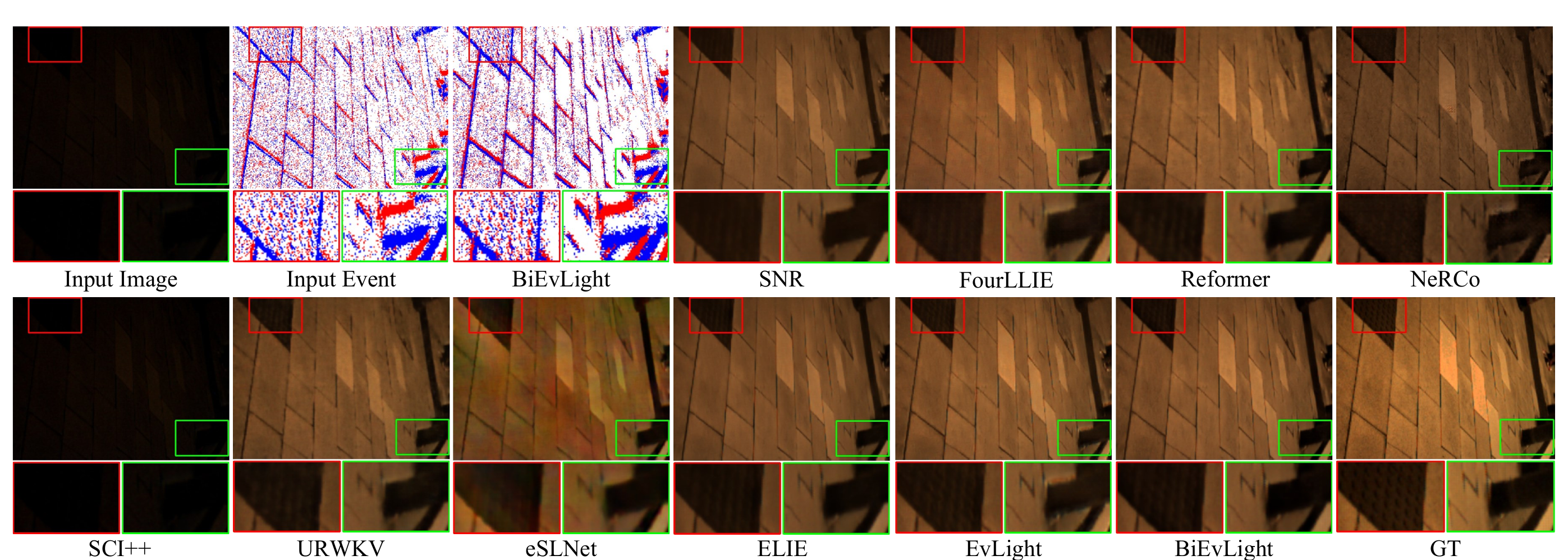}
    \caption{Qualitative results on  SDE-out dataset.}
    \label{fig7}
\centering
\end{figure*}
\begin{figure*}[!t]
\centering
    \includegraphics[width=0.98\linewidth]{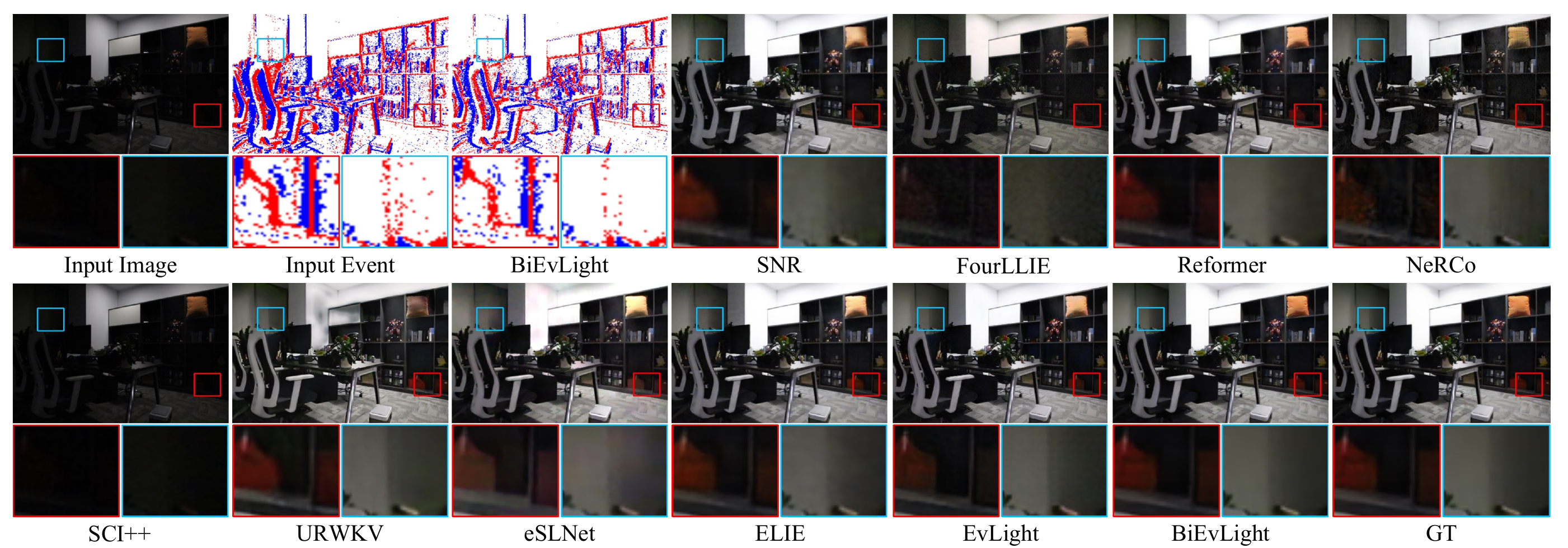}
    \caption{Qualitative results on  SDSD dataset.}
    \label{fig5}
\centering
\end{figure*}

\section{Implementation Detail}
\label{sec2}
Our models were trained using the Adam optimizer with an initial learning rate set to 2e-4. All experiments were conducted on a single NVIDIA RTX 3090 GPU, utilizing a batch size of 8. To enhance model generalization and robustness, we applied several data augmentation techniques during training. These included random cropping to a patch size of 128 × 128 pixels, and random rotations at angles of 90, 180, and 270 degrees.


\section{More  Qualitative Results}
\label{sec3}
In this section, we present additional visual examples for
various state-of-the-art models \cite{xu2022snr,wang2023fourllie,cai2023retinexformer,yang2023implicit,ma2025learning,xu2025urwkv,wang2020event,jiang2023event,liang2024towards}.

\textbf{Comparison on SDSD Dataset:}  
 Similarly, we conducted comparative experiments on the SDSD dataset \cite{wang2021seeing}, with quantitative results detailed in Tab. \ref{table1}. BiEvLight achieved optimal performance in both indoor and outdoor scenarios. Specifically, for SDSD-in and SDSD-out tasks, our method surpasses Evlight with PSNR improvements of 1.41 dB and 0.67 dB, respectively. For the PSNR* , BiEvLight demonstrates competitive advantages with improvements of 0.60 dB for SDSD-in and 0.15 dB for SDSD-out. These results comprehensively demonstrate the effectiveness of the proposed BiEvLight method.
 
 Qualitatively, as shown in Fig. \ref{fig5}, even though the event streams generated by the simulator contain only a small amount of noise, it can still degrade task performance if not addressed. As illustrated in the magnified regions, our method effectively preserves valid events while precisely removing redundant noise, and successfully recovers underexposed image details, further highlighting its efficacy.

\end{document}